\pdfoutput=1
\documentclass[10pt,twocolumn,letterpaper]{article}

\usepackage{wacv}
\usepackage{times}
\usepackage{epsfig}
\usepackage{graphicx}
\usepackage{amsmath}
\newcommand{\argmax}{\operatornamewithlimits{argmax}}
\usepackage{float}
\usepackage[utf8]{inputenc}
\usepackage{xr}
\usepackage{amssymb}
\usepackage{verbatim}
\usepackage{multirow}
\usepackage{caption}
\usepackage{subcaption}
\usepackage{makecell}
\usepackage{hyperref}

\newcolumntype{D}{ >{\centering\arraybackslash} m{0.2cm} }
\usepackage[export]{adjustbox}

\wacvfinalcopy %

\def\wacvPaperID{1187} %

\begin{document}

\title{Domain Bridge for Unpaired Image-to-Image Translation \\and Unsupervised Domain Adaptation}

\author{Fabio Pizzati$^{1,2}$, Raoul de Charette$^{2}$, Michela Zaccaria$^{3}$, Pietro Cerri$^{1}$ \\
	$^1$VisLab, $^2$Inria, $^3$University of Parma\\
	{\tt\small \{fabio.pizzati, raoul.de-charette\}@inria.fr}\\
	{\tt\small michela.zaccaria@studenti.unipr.it}\\
	{\tt\small pcerri@ambarella.com}
}
\newcommand*\rot{\rotatebox{90}}
\maketitle
\ifwacvfinal\thispagestyle{empty}\fi

\begin{abstract}
Image-to-image translation architectures may have limited effectiveness in some circumstances. For example, while generating rainy scenarios, they may fail to model typical traits of rain as water drops, and this ultimately impacts the synthetic images realism. With our method, called domain bridge, web-crawled data are exploited to reduce the domain gap, leading to the inclusion of previously ignored elements in the generated images. We make use of a network for clear to rain translation trained with the domain bridge to extend our work to Unsupervised Domain Adaptation (UDA). In that context, we introduce an online multimodal style-sampling strategy, where image translation multimodality is exploited at training time to improve performances. Finally, a novel approach for self-supervised learning is presented, and used to further align the domains. With our contributions, we simultaneously increase the realism of the generated images, while reaching on par performances with respect to the UDA state-of-the-art, with a simpler approach.

 \end{abstract}

\section{Introduction}
\begin{figure}
	\begin{subfigure}{\linewidth}
		\centering
		\includegraphics[width=0.86\linewidth]{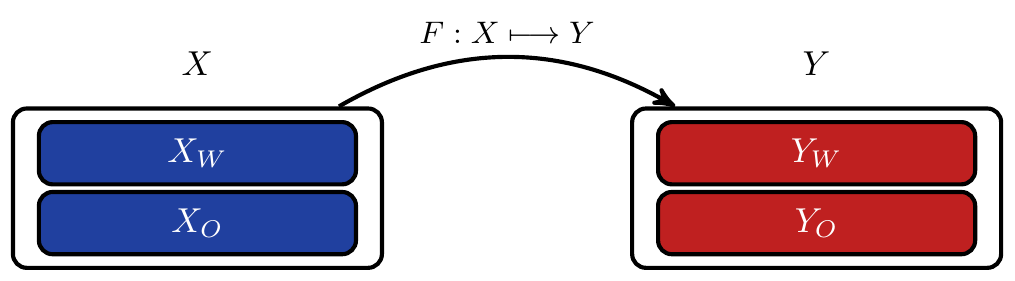}
		\caption{Naive image-to-image translation}
		\label{figure:db_naive}
	\end{subfigure}
	\begin{subfigure}{\linewidth}
		\centering
		\includegraphics[width=0.95\linewidth]{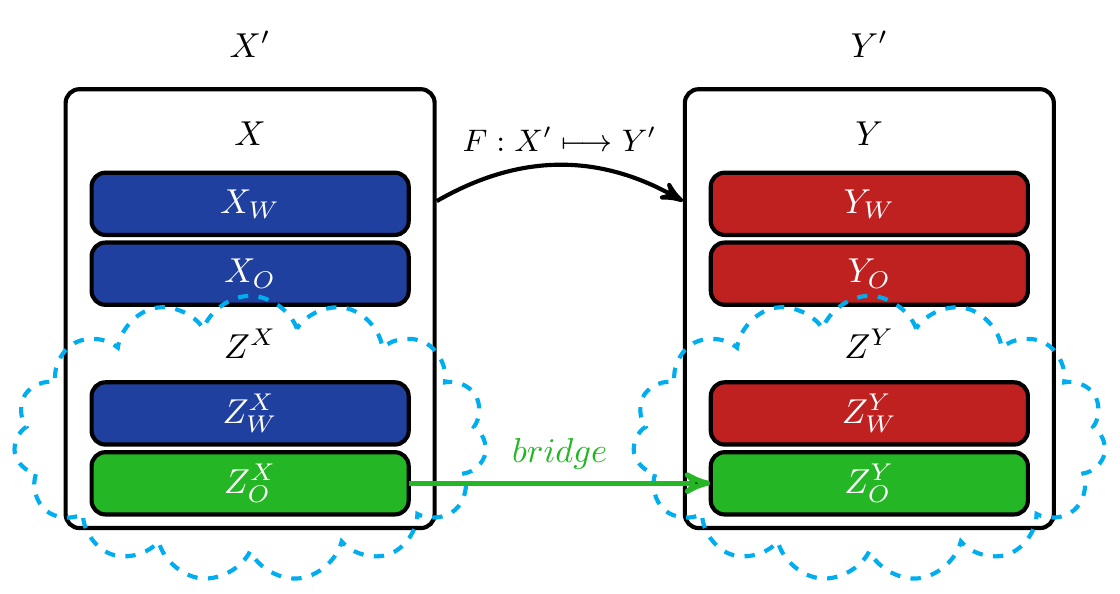}
		\caption{Domain-bridged image-to-image translation}
		\label{figure:db_bridge}
	\end{subfigure}	
	\caption{Naive image-to-image translation (Fig.~\ref{figure:db_naive}) learns the $X\rightarrow{}Y$ domain mapping. Conversely, our \textit{domain bridge} (Fig.~\ref{figure:db_bridge}) completes source and target domains with automatically retrieved web-crawled data ($Z^X$, $Z^Y$) which share common characteristics, thus easing the image-to-image translation task.}\label{figure:db} 
\end{figure}

GANs have demonstrated great ability to learn image-to-image (i2i) translation from paired~\cite{isola2017image} or unpaired images~\cite{zhu2017unpaired,huang2018multimodal} in different domains (e.g. summer/winter, clear/rainy, etc.).  
The latter relies on cycle-consistency or style/content disentanglement to learn complex mapping in an unsupervised manner, producing respectively a single translation of the source image or a multimodal translation~\cite{huang2018multimodal}.
This unsupervised i2i translation opened a wide range of applications especially for autonomous driving, for which it may be \textit{virtually} impossible to acquire the same scene in different domains (e.g. say clear/rainy) due to the dynamic nature of the scenes.

Instead, image-to-image translation can be used to generate realistic image synthesis exploitable for both domain adaptation and performance evaluation~\cite{hoffman2017cycada, li2019bidirectional}, without additional human-labeling effort.
However, state-of-the-art i2i translation fails in some situations. For example, while translating clear to rain the networks tend to change only the global appearance of the scene (wetness, puddle, etc.) ignoring essential traits as drops on the windshield or reflections. Ultimately, this greatly impacts the realism of the generated images. 

\begin{figure*}
	\includegraphics[width=\linewidth]{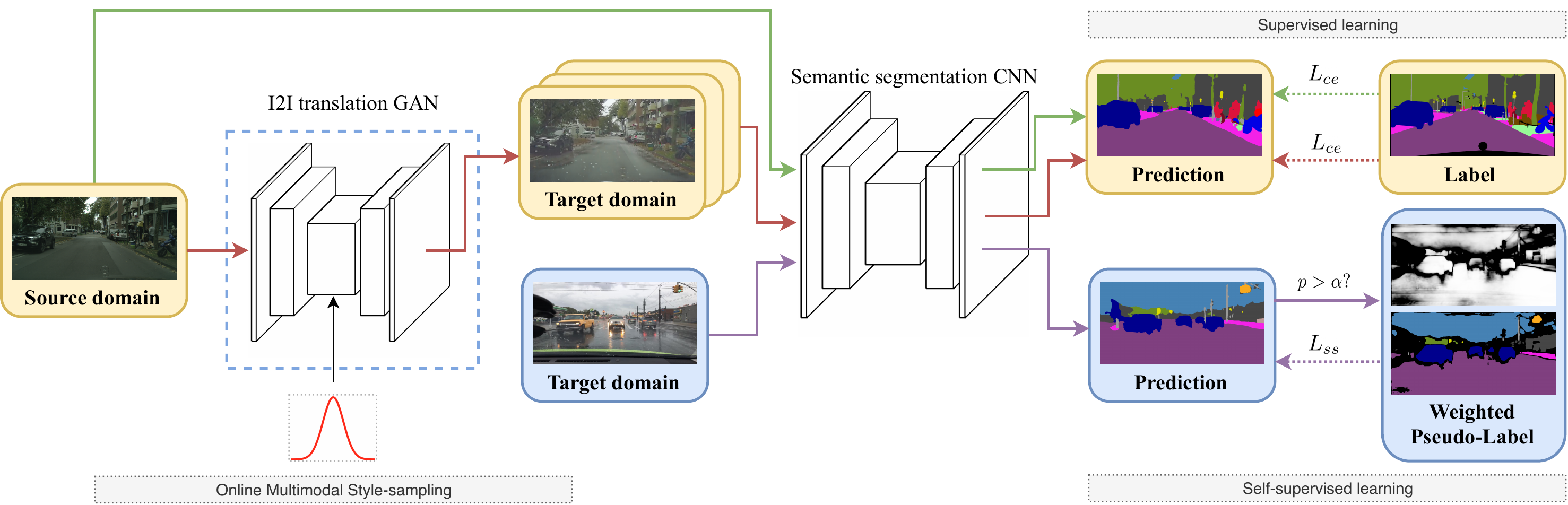}
	\caption{Overview of our pipeline for unsupervised domain adaptation. The blue dashed square means that the GAN parameters are frozen. The Image-to-image translation network is trained offline with our domain bridge strategy. Different line colors refer to different probabilities for one path to be executed. Loss functions are denoted with dotted lines.} \label{figure:systemoverview}
\end{figure*}

In this paper, we present a simple domain-bridging technique (Fig.~\ref{figure:db_bridge}) which, opposite to the standard i2i translation (Fig.~\ref{figure:db_naive}), benefits of additional sub-domains retrieved automatically from web-crawled data. 
Our method produces qualitatively significantly better results, especially when the source and target domains are known to be far since the bridge ease the learning of the mapping.
We apply our i2i methodology to the case of clear $\rightarrow$ rainy images showing that domain bridging leads the translation to preserve drops on the synthetic images, and extend our work to Unsupervised Domain Adaptation (UDA) for which we make novel contributions too (Fig.~\ref{figure:systemoverview}) and demonstrate that all together we perform on par with the most recent UDA methods while being much simpler.

\noindent{}We make three main contributions in our paper:
\begin{itemize}
\item \textbf{i2i:} we propose a novel \textit{domain-bridge} (Sec.~\ref{sec:method_bridge}) to augment automatically the source and target domains and ease i2i mapping,
\item \textbf{i2i with UDA:} \textit{online multimodal style-sampling} (Sec.~\ref{sec:method_OMS}) is applied for UDA, thus increasing the translation diversity,
\item \textbf{UDA:} we propose novel \textit{Weighted Pseudo Label} (Sec.~\ref{sec:method_wpl}) to benefit from self-supervision without the need of offline processing as for the original Pseudo Label~\cite{lee2013pseudo}.
\end{itemize}

\section{Related work}

\paragraph{Image-to-image translation.}
Early work for image-to-image translation has been done in~\cite{isola2017image}, where an adversarial-based method has been proposed. The training process required paired samples of the same scene in two different domains. In~\cite{zhu2017unpaired}, instead, cycle consistency is exploited to perform image-to-image translation on unpaired images. \cite{liu2017unsupervised} supposes the existence of a shared latent space between images in two domains, and exploits it to perform translations across both domains using a single GAN. Recently, a lot of efforts have been dedicated to achieve multimodal translation~\cite{zhu2017toward, huang2018multimodal, liu2019few}. Some others, instead, make use of additional information, such as bounding boxes~\cite{shen2019towards}, instance segmentation maps~\cite{mo2018instagan}, or semantic segmentation and depth maps~\cite{chen2019learning}, to increase the translation quality and diversity. In ~\cite{Gong_2019_CVPR}, intermediate domains are used to achieve better performances in domain adaptation.

\paragraph{Synthetic rain modeling.}
To synthesize rain on images, Garg and Nayar~\cite{garg2006photorealistic} first proposed to rely on the accurate drop oscillation and photometry modeling which has been extended in \cite{creus2013r4, tatarchuk2006artist} for rain streaks and \cite{rousseau2006realistic} for stationary drops. All assume the impractical complete 3D knowledge of the scene geometry and light sources.
Circumventing such limitations, the recent physics-based rendering \cite{halder2019physics,Hu_2019_CVPR} rely on the rain/fog layers decomposition to simulate the complete rain visual effect.
In \cite{halder2019physics}, the rain dynamics and photometry is approximated from physics models to realistically augment image or sequences with controllable amount of rain, which further allow benchmarking vision algorithms in rain.
Alternately, \cite{porav2019can} proposed to synthesize photorealistic raindrops on a screen using a simplified refraction model and approximating their dynamics with metaballs.
While priors works do produce high quality visual results, none of them face visual characteristics of rainy images, as wetness and reflections.

\paragraph{Domain adaptation for semantic segmentation.}
Most methods for domain adaptation are based on adversarial training, as it regularizes the feature extraction process, making it robust to the domain shift~\cite{fcnwild, biasetton2019unsupervised, michieli2019adversarial, sankaranarayanan2018learning, zhang2017curriculum, zhang2018fully,tsai2018learning, luo2019taking, ramirez2019learning}. Complementary approaches, instead, connect the two domains with pixel-level regularization, making use of image-to-image translation GANs~\cite{murez2018image, ramirez2018exploiting}. Some recent works combine the two to obtain better results~\cite{hoffman2017cycada, li2019bidirectional}. Others do not use adversarial training at all: for example, Zou et al. \cite{zou2018unsupervised} exploit self-supervised learning and pseudo-labels only, while~\cite{zhang2018deep} make use of mutual learning. Finally, some approaches have been recently introduced specialized on night time adaptation ~\cite{romera2019bridging} or transfer learning for adverse weather ~\cite{dai2019curriculum, sakaridis2018semantic}.

\begin{figure}
	\begin{subfigure}{0.45 \linewidth}
		\centering
		\includegraphics[width=\linewidth]{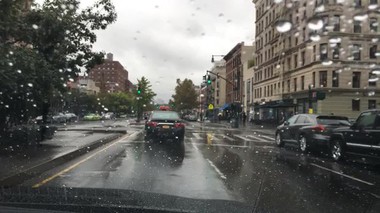}
	\end{subfigure}
	\hspace{.08\linewidth}
	\begin{subfigure}{0.45\linewidth}
		\centering	
		\includegraphics[width=\linewidth]{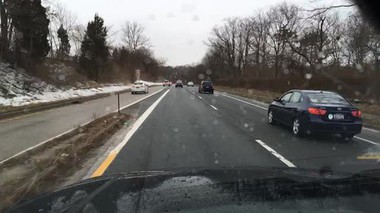}
	\end{subfigure}
	
	\caption{Rainy images from the Berkeley Deepdrive dataset~\cite{xu2017end}. Drops on the windshield or reflections help us perceive that it is raining.}\label{figure:drops}
\end{figure}

\section{Method}
Our methodology aims to translate clear images to rainy images reaching both high-qualitative images for both qualitative evaluation and usability to train semantic segmentation networks in rainy weather.
Thus, our innovations are spread between image-to-image translation (Sec.~\ref{sec:method_bridge}) and Unsupervised Domain Adaptation for semantic segmentation (Sec.~\ref{sec:method_UDA}).

\subsection{Image-to-image translation (i2i)}
Image-to-image translation GAN networks learn to approximate the mapping $F: X \mapsto Y$ using adversarial training, from two sets of representative images in each domain, denoted here $A$ and $B$. 
Each image in both sets can be interpreted as a sampling from a probability distribution $P$ associated with the image domain~\cite{liu2017unsupervised}. 
Formally, $\forall a \in A, a \sim P_X(x); \forall b \in B, b \sim P_Y(y)$.
GANs are well-known for their instability at training. 
For the latter to succeed, the network needs to be fed with representative sets of images, so that it can extract the common domain characteristics.
Even though, some domain gap may be difficult to model for the network, resulting in a loss of characteristic features of the target domain. 
This may be caused by a significant domain shift or by the lack of data. 

Some minor image details still have a significant perceptual impact. This is the case for rain images, where even a few drops on the lens play an important role in \textit{sensing} the rain, as it can be seen in Fig.~\ref{figure:drops}.
State-of-the-art networks may ignore some fundamental elements as drops, lens artifacts or reflections, and this ultimately impacts the realism of generated images.
We argue that some characteristics changes (drops, artifacts, etc.) are ignored because they are relatively minor compared to other characteristics changes (e.g. wetness, puddles, etc.), and demonstrate the training may benefit from bridging to ease domain mapping.

\paragraph{Domain bridging.}
\label{sec:method_bridge}

Studying the specific case of adverse weather conditions, it is possible to formalize a generic domain $K$ as the union of finer-grained domains, such as $K = \{K_W, K_O\}$. In it, $K_W$ represents the sub-domains typical of weather, e.g. the presence of precipitations, road wetness, and many more. $K_O$, instead, is composed of sub-domains unrelated to the weather. Examples are the scenario, the city, and the illumination. Thus, it is possible to represent $X$ and $Y$ as
\begin{equation}
\begin{split}
X = \{X_W, X_O\}\ ,\\\
Y = \{Y_W, Y_O\}\ .
\end{split}
\end{equation}
Generally, only the joint probability distribution $P_K(k) = P_{K_W, K_O}(k), k \in K$ is estimable, as we have no knowledge about the marginal probability distrubtions $P_{K_W}(k)$, $P_{K_O}(k)$. 

On one hand, we hypothesize that it is possible to obtain a more stable image-to-image translation if the differences between the two datasets are minimized. On the other, we have to obtain a GAN that produces an effective transformation, so it is necessary to model correctly all relevant features of adverse weather. 
To simultaneously reach both objectives, images collected from web-crawled videos are added to the $A$ and $B$ datasets, obtaining two new training sets $A'$ and $B'$, with respective domains $X'$ and $Y'$, which aims as bridging the gap between the initial domain $X$ and $Y$. 
This is illustrated in Fig.~\ref{figure:db}.\\
Our intuition is that adding samples with reasonable criteria will reduce the Kullback-Leibler divergence between the probability distributions $P_{X'_O}(x)$ and $P_{Y'_O}(y)$, with respect to $KL(P_{X_O}(x),P_{Y_O}(y))$. As a consequence, the translation model will be more focused on weather-related characteristics and more stable during training.

Once the main hypotheses are formalized, the approach on how to select new images is needed. Let $Z^{x}$ and $Z^{y}$ be two images sets. As before, we have 
\begin{equation}
\begin{split}
Z^x = \{Z^x_W, Z^x_O\}\ ,\\
Z^y = \{Z^y_W, Z^y_O\} .
\end{split}
\end{equation}
We choose $Z^{x}$ and $Z^{y}$ in order to have 
\begin{equation}
\label{equation:card}
\begin{split}
max \ \vert Z^x_W \cap X_W\vert\ ,\\\
max \ \vert Z^y_W \cap Y_W \vert\ , \\
max \ \vert Z^x_O \cap Z^y_O \vert\ ,\\
\end{split}
\end{equation}
where $\vert\cdot\vert$ is the set cardinality. Hence, it is possible to identify two image sets $C$ and $D$ such as $\forall c \in C, c \sim P_{Z^x}(z); \forall d \in D, d \sim P_{Z^y}(z)$. 

It is now possible to train the image-to-image translation network on $A'$ and $B'$ defined as:
\begin{equation}
\begin{split}
A' = A \cup C\ , \\
B' = B \cup D\ .
\end{split}
\end{equation}
Adding new images, the differences in the global appearance of the two domains is minimized, while the weather-related domain shift remains constant.

In other words, our approach consists in selecting new image samples, with weather conditions corresponding to those in the original dataset, and join them to the existing data. The newly-added images are required to share some domains unrelated to the weather. Retrieving images from the same location and with the same setup ensures that.

In practice, we retrieved these additional samples from web-crawled videos using domain-related keywords search (details in Sec.~\ref{sec:exp_datasets}).
The same bridging can be applied automatically to other domain shifts, though as the domain differences become less semantically evident, human expertise may be required to properly select the $C$ and $D$ datasets.
\\

We use MUNIT~\cite{huang2018multimodal} as backbone for our image-to-image translation network, as it allows disentanglement of style and content, which will be of high interest for us.

\begin{figure}
	\centering
	\resizebox{\linewidth}{!}{\begin{tabular}{D c c}
			\rotatebox{90}{Source} 
			& \includegraphics[width=128px, valign=m]{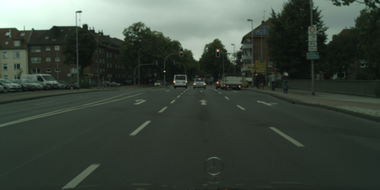} 
			& \includegraphics[width=128px, valign=m]{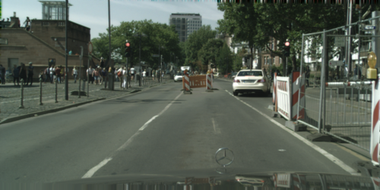} \\
			\noalign{\smallskip} \rotatebox{90}{Style 1} 
			& \includegraphics[width=128px, valign=m]{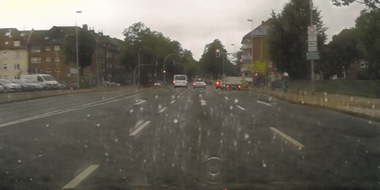} 
			& \includegraphics[width=128px, valign=m]{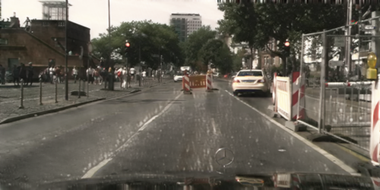}  \\
			\noalign{\smallskip} \rotatebox{90}{Style 2} 
			& \includegraphics[width=128px, valign=m]{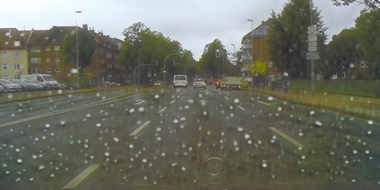} 
			& \includegraphics[width=128px, valign=m]{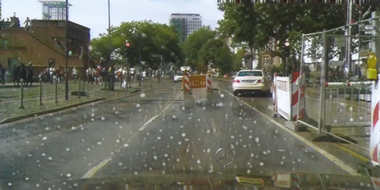}  \\
			\noalign{\smallskip} \rotatebox{90}{Style 3} 
			& \includegraphics[width=128px, valign=m]{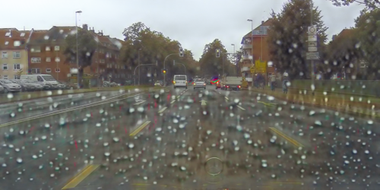} 
			& \includegraphics[width=128px, valign=m]{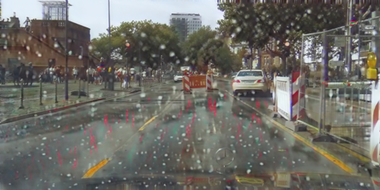} \\
			\vspace{20px}
			
	\end{tabular}}
	\caption{Examples of multimodal style-sampling from our Domain-bridged i2i (Sec.~\ref{sec:method_bridge}). Note the consistency of style regardless of the source image.}\label{figure:multimodal}
\end{figure}
\subsection{Unsupervised Domain Adaptation (UDA)}
\label{sec:method_UDA}
Similar to previous works~\cite{hoffman2017cycada, li2019bidirectional}, we use our i2i network with domain-bridge to translate images from pre-labeled \textit{clear weather} datasets and learn semantic segmentation in rain in an unsupervised fashion. 
We follow the standard UDA practice which is to alternately train in a supervised manner from source (\textit{clear}) images with labels and train in a self-supervised manner from target (\textit{rain}) images without labels. 
Our entire UDA methodology (depicted in Fig.~\ref{figure:systemoverview}) brings two novels contributions: a) we use multi-modal clear$\rightarrow$rain translations as additional supervised learning, b) we introduce Weighted Pseudo Label - a differentiable extension of Pseudo Label~\cite{lee2013pseudo} - to align source and target without any offline process.

\paragraph{Online Multimodal Style-sampling (OMS).}\label{sec:method_OMS}

The standard strategy for UDA with i2i networks is to learn from the offline translation of the whole dataset~\cite{hoffman2017cycada, li2019bidirectional}. 

We instead propose to use the multimodal capacity of MUNIT i2i to generate multiple target styles (i.e. \textit{rain appearances}) for each input image. Styles are sampled during training time.
In this way, even if the source image content remains unaltered, it will be possible for the segmentation network to learn different representations of the same scene in the target domain, ultimately leading towards wider diversity and thus more robust detection.
Different styles for the same image modify, among other factors, the position and size of drops on the windshield, and the intensity of reflections. 
This is visible in Fig.~\ref{figure:multimodal} showing three arbitrarily sampled styles, where Style~3 consistently produces images that resemble heavy rain.

\paragraph{Weighted Pseudo-Labels (WPL).}
\label{sec:method_wpl}
Pseudo Label~\cite{lee2013pseudo} was proposed as a self-supervised loss to further align source and target distributions.
The principle is to self-train a network on target (here, \textit{rain}) whenever the prediction confidence is above some threshold, thus reinforcing the network beliefs. 

\noindent{}Most often for UDA, thresholds are calculated offline as the median per-class confidence dataset-wise~\cite{li2019bidirectional,zou2018unsupervised}.  
This requires storing all predictions for the whole dataset, which is cumbersome. 
To overcome this, thresholds may be estimated online image-wise or batch-wise~\cite{iscen2019label}.
It has to be noted that thresholds are critical since pseudo-labeling can harm global performances if thresholds were underestimated\footnote{In such case, the ratio of wrong pixels over pseudo-labeled pixels will be too high and lead to incorrect self-supervision.} or have limited impact if overestimated~\cite{lee2013pseudo}. 

We instead propose Weighted Pseudo-Labels (WPL) which estimates a global threshold $\alpha$ within the network optimization process. 
The general principle of WPL is to weight the self-supervised cross-entropy using learned threshold $\alpha$, thus acting as continuous pseudo-labeling.
Not only WPL does not require offline processing, but it is aware of the global network confidence thus leading to better results.
In detail, let $x$ be an input image and $\hat{x}_u$ the pseudo label of pixel $u$, such that
\begin{equation}
\label{equation:pseudolabel}
\hat{x}_u = \begin{cases} 
	\argmax_{q} f_u(x) & \text{if }\max(f_u(x)) \geq \alpha  \\
	\text{None} & \text{otherwise}
	\end{cases}\ ,
\end{equation}
where $f_u(x)$ refers to the class probabilities of $u$ predicted by the network $f$, and $\argmax_{q}$ is the best class prediction. 
In its original implementation~\cite{lee2013pseudo}, $\hat{x}_u$ is directly used to weight the cross-entropy self-supervision. 
Instead, we weight this with a weight matrix $W$ of the same size than $x$:
\begin{equation}
\label{equation:weight}
	w_u = \begin{cases} 
	\frac{\max(f_u(x)) - \alpha}{1 - \alpha} & \text{if }\max(f_u(x)) \geq \alpha \\
	0 & \text{otherwise}
	\end{cases}\ ,
\end{equation}
The complete loss for WPL is thus defined as the weighted sum of cross-entropy loss $L_{w}$ and a balancing loss $L_b$:
\begin{equation}
\label{loss-ss}
L_{ss} = \sigma L_{w} + \gamma L_b\ ,
\end{equation}
where $\sigma$ and $\gamma$ are loss weights and cross-entropy loss is:
\begin{equation}
L_w = - \sum_{q \in Q}w_u\hat{x}_{u,q}log(p_{u})\ .
\end{equation}
$\hat{x_u}$ is the one-hot encoding of pseudo-label as in Eq.~\ref{equation:pseudolabel} and $Q$ is the set of classes. 
In this way, predictions where the network is uncertain are weighted less in the network pseudo-label based training. 
To avoid that the self-supervised contribution remains set to zero by the optimizer, $L_b$ is required as a balancing loss:
\begin{equation}
L_b = log^2(1 - \alpha)\ .
\end{equation}

\begin{figure}
	\centering
	\includegraphics[width=\linewidth]{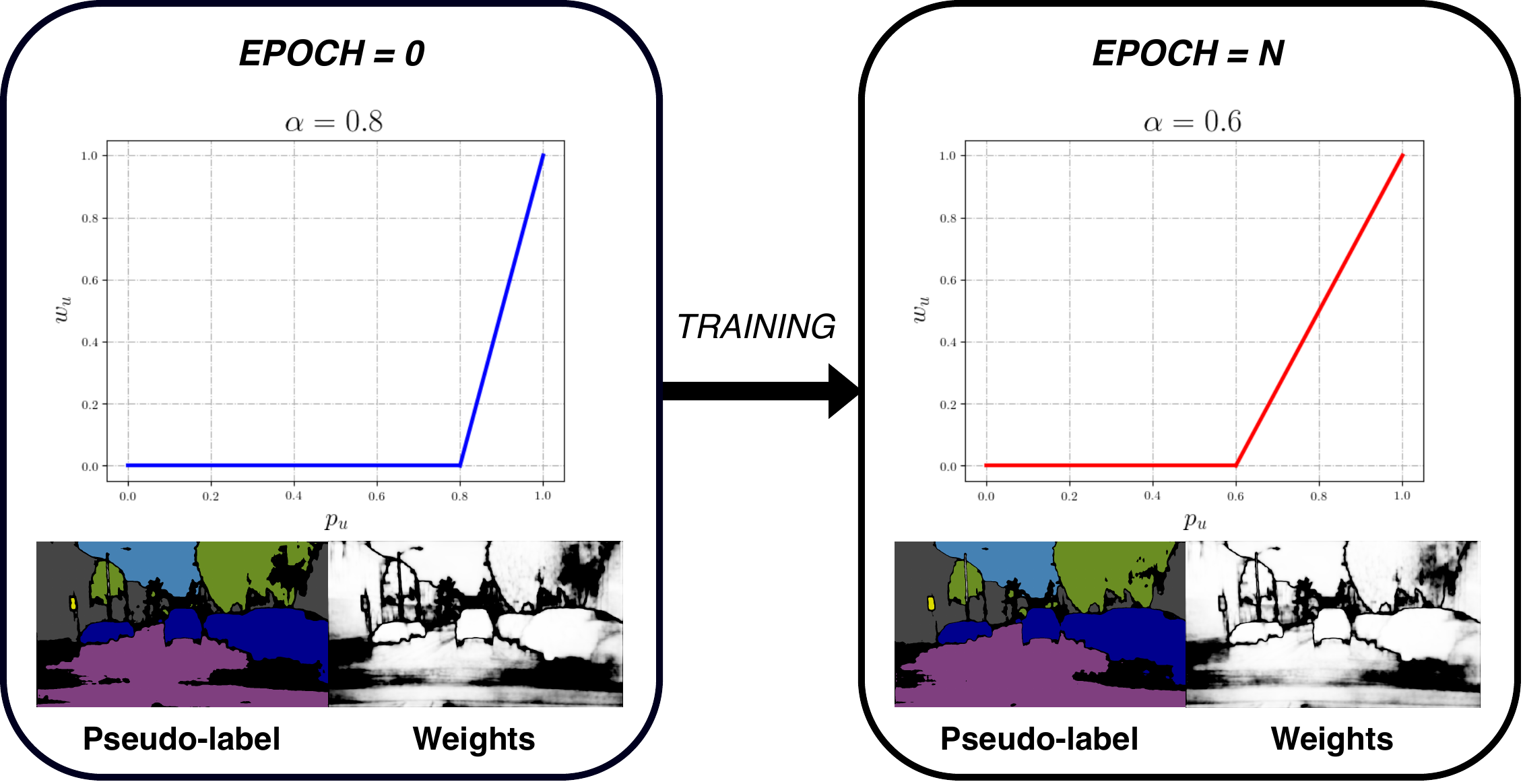}
	\caption{Analysis of the effect of $\alpha$ optimization. During training the Weighted Pseudo Label expands from high confidence pixels only (left) to lower ones (right).} \label{figure:pl}
\end{figure}
The optimization of $\alpha$ leads to a pseudo-label expansion within the training process. 
Fig.~\ref{figure:pl} is an illustration of the growing process during training.
For the first iteration (Fig.~\ref{figure:pl}, left), the $L_w$ term prevails over $L_b$, pushing $\alpha$ towards 1 thus including in the pseudo-label only pixels with high confidence.
With the minimization of $L_w$ (Fig.~\ref{figure:pl}, right), $L_b$ becomes gradually more important, leading the network to simultaneously include lower confidence pixels inside the pseudo-label, and increasing the informative potential of higher-confidence labels. 
Note that for numerical stability, we assume $\alpha = \text{sigmoid}(\beta)$ and estimate $\beta$.

\paragraph{Losses.}
To balance the self-supervised WPL contribution with the supervised learning in segmentation, we employ a probability-based approach where pseudo-label is applied only if a uniformly sampled variable $p_{pl} \in U(0,1)$ is above a predefined threshold $p_{tp}$.
Hence, the complete UDA loss function is:
\begin{equation}
\begin{split}
L(x_a, \hat{x}_a, x_b) = L_{ce}(f(x_a), \hat{x}_a) \\ + L_{ss}(f(x_b))[p_{pl} > p_{tp}]
\end{split}
\label{eq:UDA_loss_source}
\end{equation}
if we train on source data + target, and
\begin{equation}
\begin{split}
L(x_a, \hat{x}_a, x_b) = L_{ce}(f(g(x_a)), \hat{x}_a) \\+ L_{ss}(f(x_b))[p_{pl} > p_{tp}] 
\end{split}
\label{eq:UDA_loss_target}
\end{equation}
if we train on translated images + target. In Eq.~\ref{eq:UDA_loss_source} and \ref{eq:UDA_loss_target}, $x_a \in A$ is source image with label $\hat{x}_a$, $x_b \in B$ target image, $L_{ce}$ the cross-entropy loss, $f$ our segmentation network, $g$ our bridged-GAN, and $[\cdot]$ are the Iverson brackets.

\section{Experiments}
We now evaluate the performance of both our i2i proposal (Sec.~\ref{sec:exp_i2i}) and our UDA proposal (Sec.~\ref{sec:exp_UDA}) on the clear$\rightarrow$rain problem using clear/rain datasets recorded with different setups.

\subsection{Experimental settings}
\subsubsection{Datasets}
\label{sec:exp_datasets}
For i2i and UDA, we use the german dataset Cityscapes~\cite{cityscapes} as source (\textit{clear}), and a subset of the American Berkeley DeepDrive~\cite{xu2017end} (BDD) as target (\textit{rain}). The bridge dataset, only used for the i2i, is a collection of Youtube videos. We now detail each dataset.

\paragraph{Cityscapes.} We train on Cityscapes training set with 2975 pixel-wise annotated images, and evaluate on their validation set with 500 images. While we train on crops, we evaluate on full-size images, i.e. $2048\times1024$. The \textit{trainExtra} set, with 19997 images, is also included in the domain bridge to further reduce the domain shift.

\paragraph{BDD-rainy.}
We use the coarse weather annotations of BDD together with daylight annotation to obtain a subset we call BDD-rainy (i.e. rainy+daylight), i.e. $1280\times720$.
For training the rainy+daylight is extracted from the 100k split, while for validation only the 10k split is used. Obviously, duplicates present in both splits are removed.
It has to be noted that, while daylight annotation is accurate, weather annotation is approximate and "rainy" images may either be taken during or \textit{after} a rain event, thus with or without drops on the windshield. This further increases complexity.

\paragraph{Bridge dataset.}
\begin{figure}
	\centering
	\scriptsize
	\setlength{\tabcolsep}{0.01\linewidth}
	\renewcommand{\arraystretch}{0.5}
	\begin{tabular}{D c c c}
		\rotatebox{90}{Clear weather} & \includegraphics[width=0.3\linewidth, valign=m]{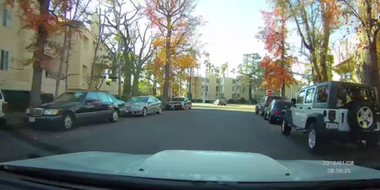} & \includegraphics[width=0.3\linewidth, valign=m]{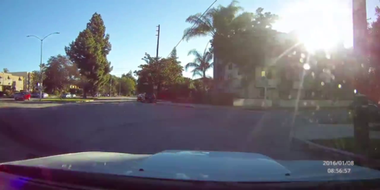} & \includegraphics[width=0.3\linewidth, valign=m]{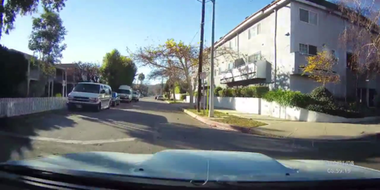} \\
		\noalign{\smallskip} \rotatebox{90}{Rain} & \includegraphics[width=0.3\linewidth, valign=m]{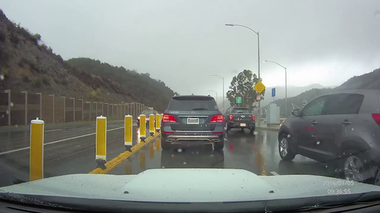} & \includegraphics[width=0.3\linewidth, valign=m]{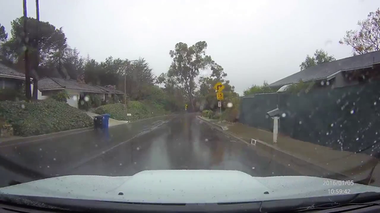}& \includegraphics[width=0.3\linewidth, valign=m]{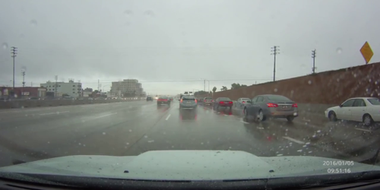}\\
	\end{tabular}
	\caption{Samples from the bridge dataset in different weather conditions. Note that the acquisition setup (camera position, optics, etc.) remains unaltered.}\label{figure:sampleyoutube}
\end{figure}

5 sequences (1280 $\times$ 720) were extracted from a single Youtube channel with keywords "driving" (2 videos) for clear weather and "driving rain"/"driving heavy rain" (2/1 video) for rainy scenarios.
The choice of using videos from a unique channel further reduces domain gaps, ensuring the same acquisition setup. Some samples are shown in Fig.~\ref{figure:sampleyoutube}.
Also to maximize image diversity, videos are uniformly sub-sampled into 2x6026 clear weather images and 3x9294 rainy images, leading to a total of 39934 frames.

 \begin{table}
 	\begin{subfigure}{.45\linewidth}
 		\centering
 		\scriptsize
 		\begin{tabular}{c|c|c}
 			\textbf{Network} & \textbf{LPIPS} & \textbf{IS} \\
 			\hline Real images & 0.7137 & - \\
 			\hline\hline CycleGAN \cite{zhu2017unpaired} & 0.1146 & 1.15\\
 			\hline MUNIT \cite{huang2018multimodal} & \textbf{0.3534} & \textbf{1.92}\\
 			\hline MUNIT-Bridged & 0.2055 & 1.69\\
 		\end{tabular}
 		\caption{GAN metrics}\label{table:ganmetrics}
 	\end{subfigure}\hspace{.08\linewidth}
 	\begin{subfigure}{.45\linewidth}
 		\centering
 		\scriptsize
 		\begin{tabular}{c|c}
 			\textbf{Network} & \textbf{mIoU} \\
 			\hline Baseline & 31.67 \\
 			\hline CycleGAN \cite{zhu2017unpaired} & 35.09\\
 			\hline MUNIT \cite{huang2018multimodal} & 20.78\\
 			\hline MUNIT-Bridged & \textbf{35.18}\\
 			
 		\end{tabular}
 		\caption{Semantic Segmentation}\label{table:semseg}
 	\end{subfigure}
 	
 	\caption{Quantitative evaluation of translated image realism, diversity, and semantic segmentation effectiveness.}\label{table:lpips}
 \end{table}
 
\subsubsection{Networks details}
\paragraph{Image-to-image translation}\label{paragraph:i2isettings}
During training, the images are downsampled to be 720 pixels in height, and cropped to 480 $\times$ 480 resolution. The network is trained for 200k iterations, with batch size 1. Adam is used as optimizer, with learning rate 1e-4, $\beta_1=0.5$, $\beta_2=0.999$.

\paragraph{Segmentation.}\label{paragraph:segmentation}
We use Light-weight Refinenet~\cite{Nekrasov2018LightWeightRF} with Resnet-101 as backbone, pretrained on the full-size Cityscapes dataset.
The refining is achieved by training for 100 epochs on 512 $\times$ 512 crops, after downscaling images to $1024\times512$ for GPU memory constraints.
We employ data augmentation for the training process, with random rescaling between a factor 0.5 and 2, and random horizontal flipping. The batch size used is 6. We use the SGD optimizer with different learning rates for the encoder (1e-4) and the decoder (1e-3). The momentum is set to 0.9, and the learning rate is divided by 2 every 33 epochs. When pseudo-labels are added to the training, we further refine the network for 70 additional epochs, with constant learning rate divided by 10 with respect to the initial values. The $\alpha$ parameter is initialized to 0.8 and estimated by SGD as well, with learning rate 0.01 and momentum 0.9. 
 
 \begin{figure*}
 	\centering
 	\resizebox{\linewidth}{!}{\begin{tabular}{D D c c c c c c c }
 			
 			\noalign{\smallskip} 
 			& \rotatebox{90}{Image} 
 			& \includegraphics[width=128px, valign=m]{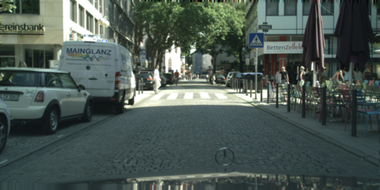} 
 			& \includegraphics[width=128px, valign=m]{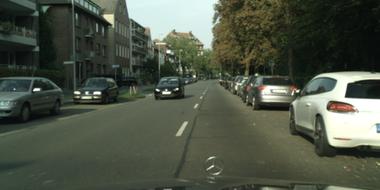} 
 			& \includegraphics[width=128px, valign=m]{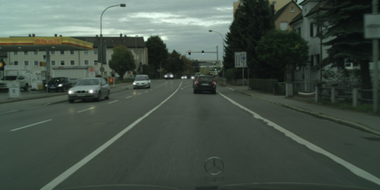} 
 			& \includegraphics[width=128px, valign=m]{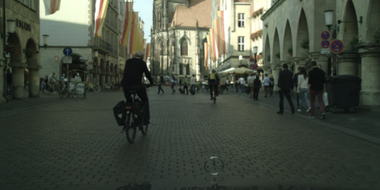} 
 			& \includegraphics[width=128px, valign=m]{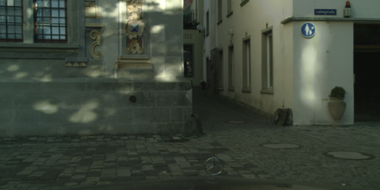} 
 			& \includegraphics[width=128px, valign=m]{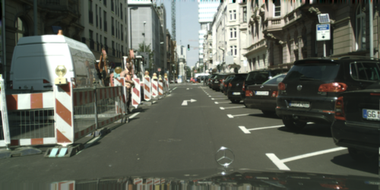} 
 			& \includegraphics[width=128px, valign=m]{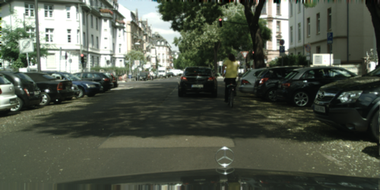}  \\
 			\noalign{\smallskip} 
 			& \rotatebox{90}{Cyclegan \cite{zhu2017unpaired}} 
 			& \includegraphics[width=128px, valign=m]{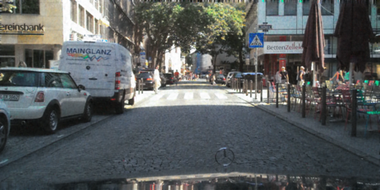} 
 			& \includegraphics[width=128px, valign=m]{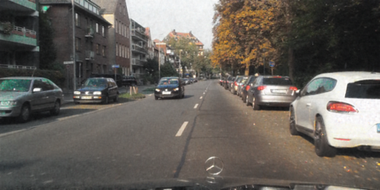} 
 			& \includegraphics[width=128px, valign=m]{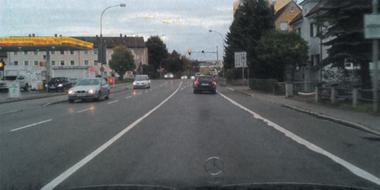} 
 			& \includegraphics[width=128px, valign=m]{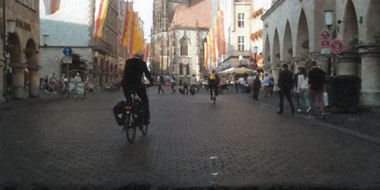} 
 			& \includegraphics[width=128px, valign=m]{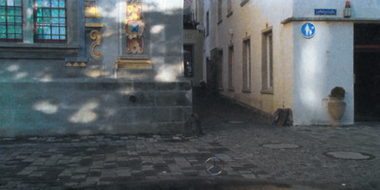} 
 			& \includegraphics[width=128px, valign=m]{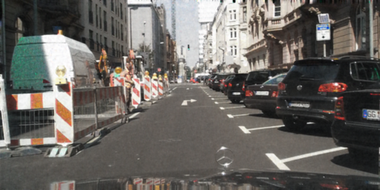} 
 			& \includegraphics[width=128px, valign=m]{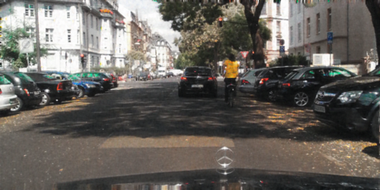}  \\
 			\noalign{\smallskip} 
 			& \rotatebox{90}{MUNIT \cite{huang2018multimodal}} 
 			& \includegraphics[width=128px, valign=m]{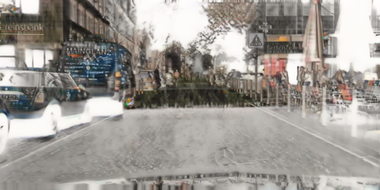} 
 			& \includegraphics[width=128px, valign=m]{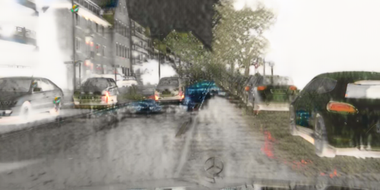} 
 			& \includegraphics[width=128px, valign=m]{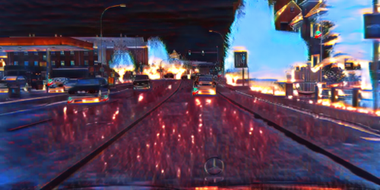} 
 			& \includegraphics[width=128px, valign=m]{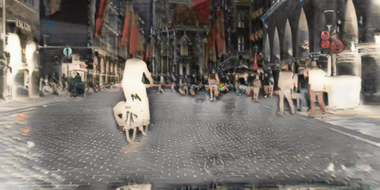} 
 			& \includegraphics[width=128px, valign=m]{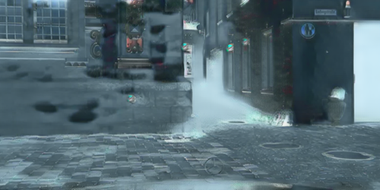} 
 			& \includegraphics[width=128px, valign=m]{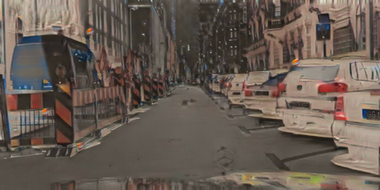} 
 			& \includegraphics[width=128px, valign=m]{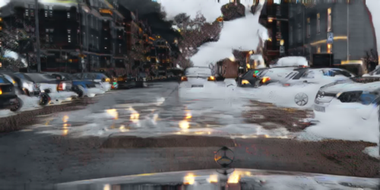}  \\
 			\noalign{\smallskip} \rotatebox{90}{\textbf{Ours}} 
 			& \rotatebox{90}{MUNIT-Bridged} 
 			& \includegraphics[width=128px, valign=m]{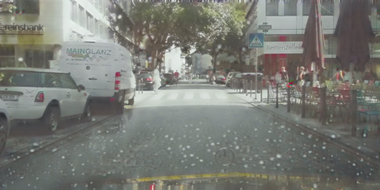} 
 			& \includegraphics[width=128px, valign=m]{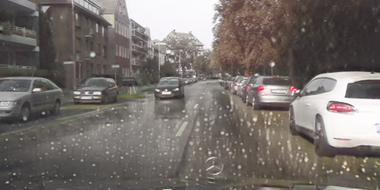} 
 			& \includegraphics[width=128px, valign=m]{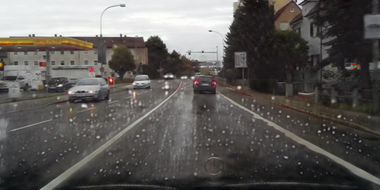} 
 			& \includegraphics[width=128px, valign=m]{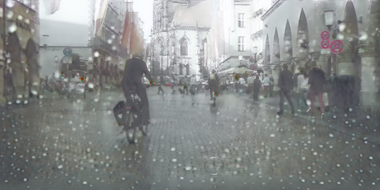} 
 			& \includegraphics[width=128px, valign=m]{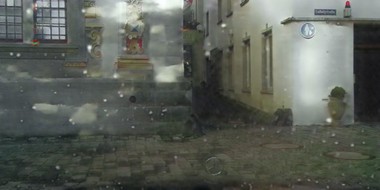} 
 			& \includegraphics[width=128px, valign=m]{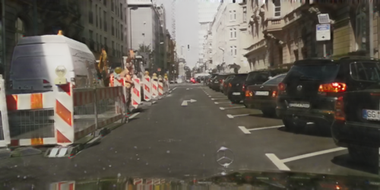} 
 			& \includegraphics[width=128px, valign=m]{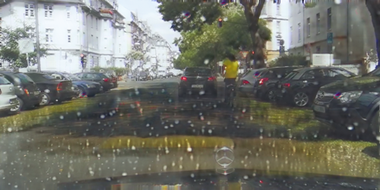}  \\
 			\noalign{\smallskip} \rotatebox{90}{\textbf{Ours}} 
 			& \rotatebox{90}{MUNIT-Bridged} 
 			& \includegraphics[width=128px, valign=m]{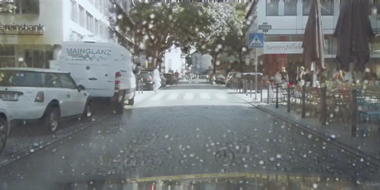} 
 			& \includegraphics[width=128px, valign=m]{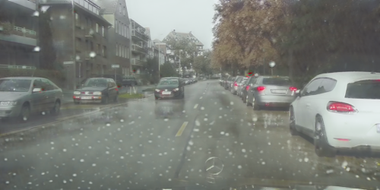} 
 			& \includegraphics[width=128px, valign=m]{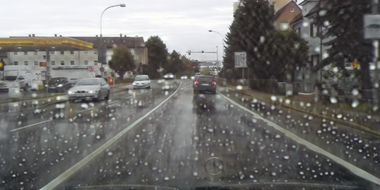} 
 			& \includegraphics[width=128px, valign=m]{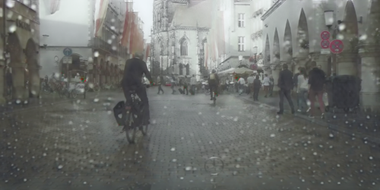} 
 			& \includegraphics[width=128px, valign=m]{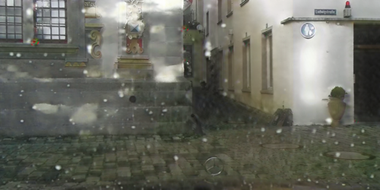} 
 			& \includegraphics[width=128px, valign=m]{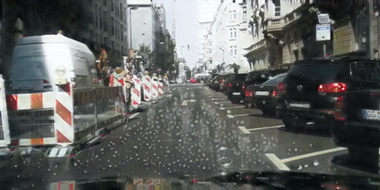} 
 			& \includegraphics[width=128px, valign=m]{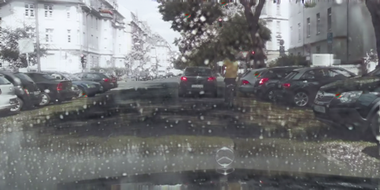}  \\
 			\noalign{\smallskip} \rotatebox{90}{\textbf{Ours}} 
 			& \rotatebox{90}{MUNIT-Bridged} 
 			& \includegraphics[width=128px, valign=m]{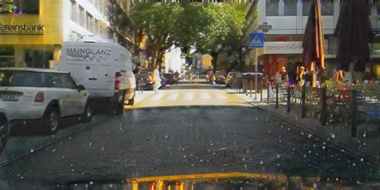} 
 			& \includegraphics[width=128px, valign=m]{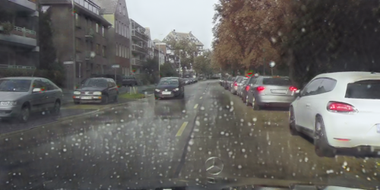} 
 			& \includegraphics[width=128px, valign=m]{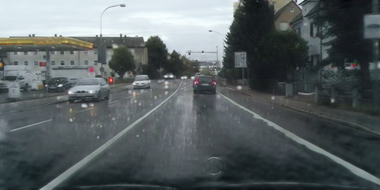} 
 			& \includegraphics[width=128px, valign=m]{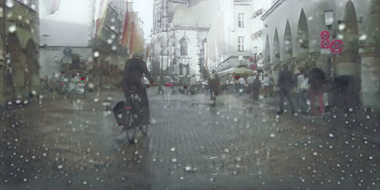} 
 			& \includegraphics[width=128px, valign=m]{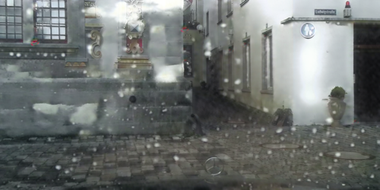} 
 			& \includegraphics[width=128px, valign=m]{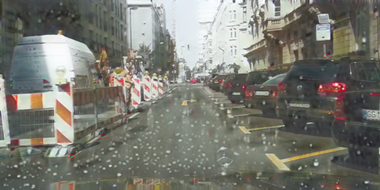} 
 			& \includegraphics[width=128px, valign=m]{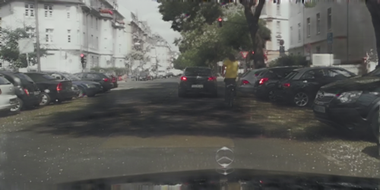}  \\
 			
 	\end{tabular}}
 	 	\caption{Qualitative comparison between state-of-the-art architectures for i2i in the clear $\rightarrow$ rain transformation. }\label{figure:samefig}

 \end{figure*}
 \begin{figure*}
 	\centering
 	\resizebox{\linewidth}{!}{\begin{tabular}{D c c c c c c c }
 			
 			\rotatebox{90}{Image} 
 			& \includegraphics[width=128px, valign=m]{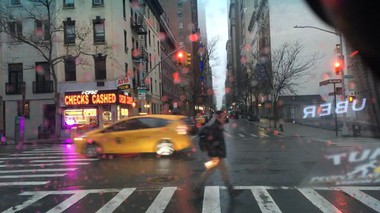} 
 			& \includegraphics[width=128px, valign=m]{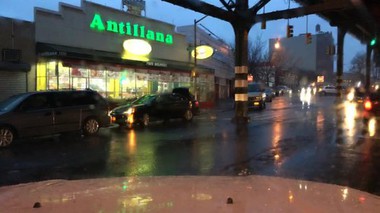} 
 			& \includegraphics[width=128px, valign=m]{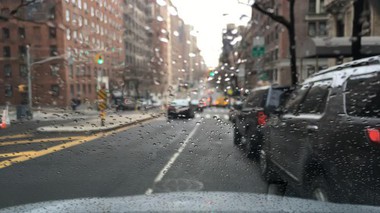} 
 			& \includegraphics[width=128px, valign=m]{figures/bdd_image_4} 
 			& \includegraphics[width=128px, valign=m]{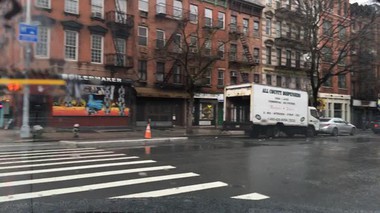} 
 			& \includegraphics[width=128px, valign=m]{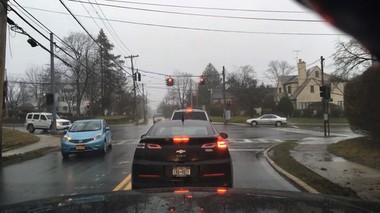} 
 			& \includegraphics[width=128px, height=72px, valign=m]{figures/bdd_image_7}  \\
 			\noalign{\smallskip} \rotatebox{90}{Ground truth} 
 			& \includegraphics[width=128px, valign=m]{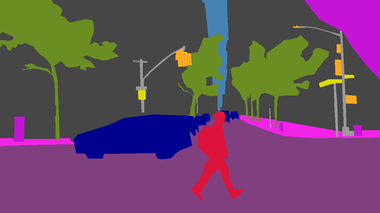} 
 			& \includegraphics[width=128px, valign=m]{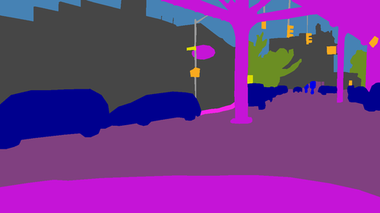} 
 			& \includegraphics[width=128px, valign=m]{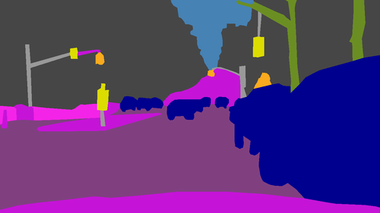} 
 			& \includegraphics[width=128px, valign=m]{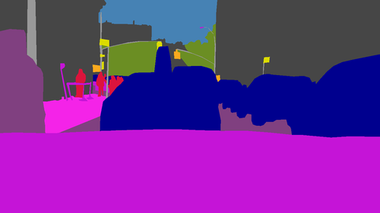} 
 			& \includegraphics[width=128px, valign=m]{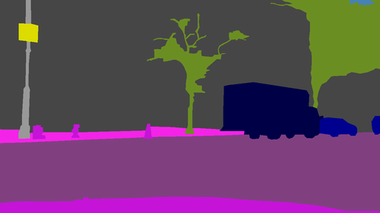} 
 			& \includegraphics[width=128px, valign=m]{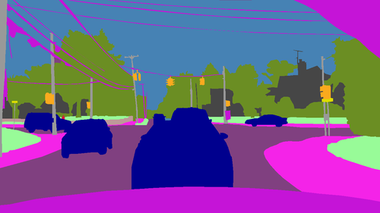} 
 			& \includegraphics[width=128px, valign=m]{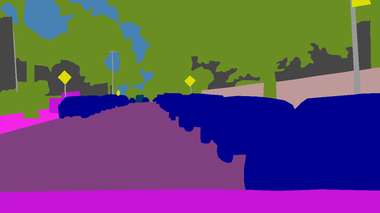}  \\
 			\noalign{\smallskip} \rotatebox{90}{Baseline} 
 			& \includegraphics[width=128px, valign=m]{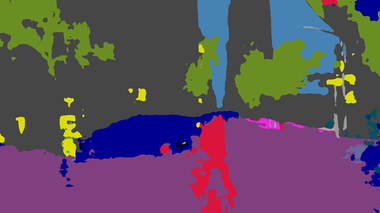} 
 			& \includegraphics[width=128px, valign=m]{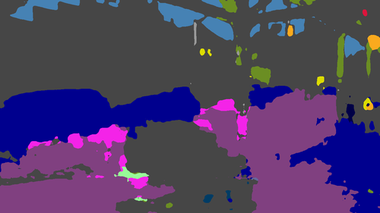} 
 			& \includegraphics[width=128px, valign=m]{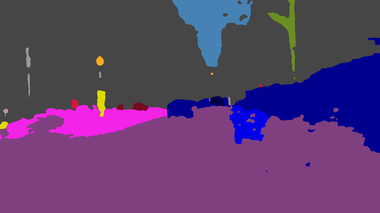} 
 			& \includegraphics[width=128px, valign=m]{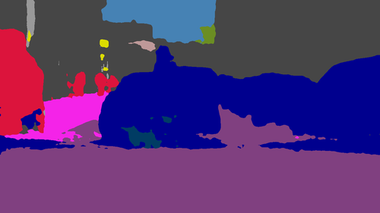} 
 			& \includegraphics[width=128px, valign=m]{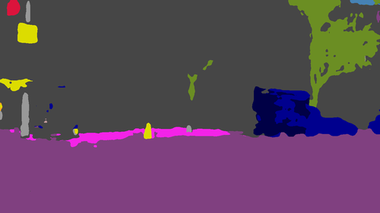} 
 			& \includegraphics[width=128px, valign=m]{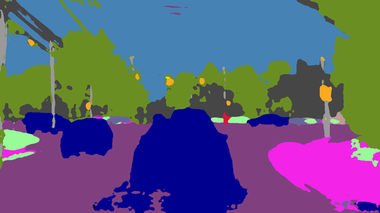} 
 			& \includegraphics[width=128px, valign=m]{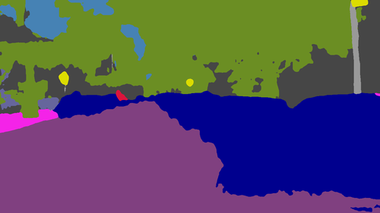}  \\
 			\noalign{\smallskip} \rotatebox{90}{AdaptSegNet \cite{tsai2018learning}} 
 			& \includegraphics[width=128px, valign=m]{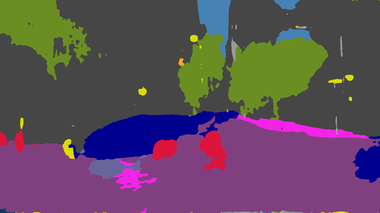} 
 			& \includegraphics[width=128px, valign=m]{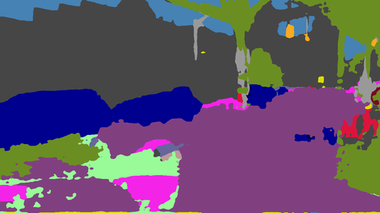} 
 			& \includegraphics[width=128px, valign=m]{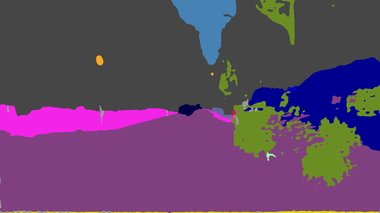} 
 			& \includegraphics[width=128px, valign=m]{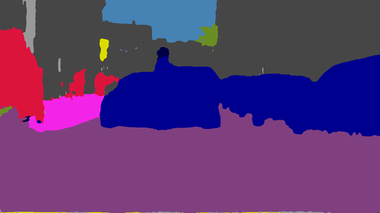} 
 			& \includegraphics[width=128px, valign=m]{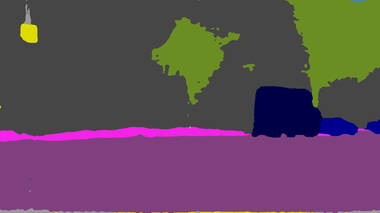} 
 			& \includegraphics[width=128px, valign=m]{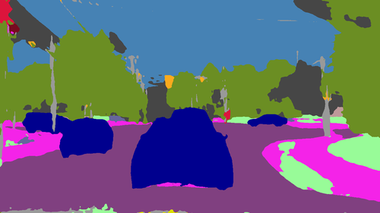} 
 			& \includegraphics[width=128px, valign=m]{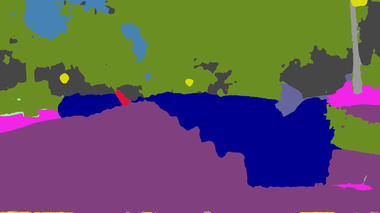}  \\
 			\noalign{\smallskip} \rotatebox{90}{BDL \cite{li2019bidirectional}} 
 			& \includegraphics[width=128px, valign=m]{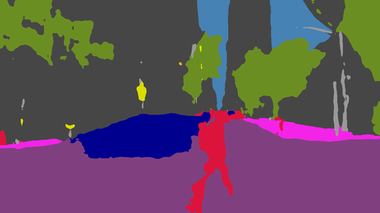} 
 			& \includegraphics[width=128px, valign=m]{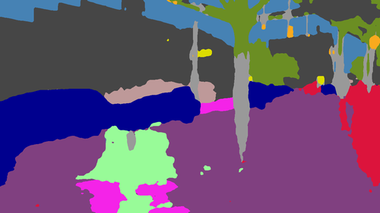} 
 			& \includegraphics[width=128px, valign=m]{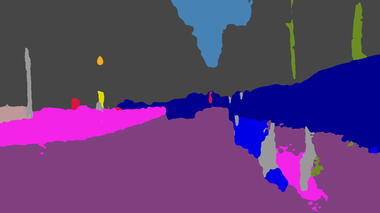} 
 			& \includegraphics[width=128px, valign=m]{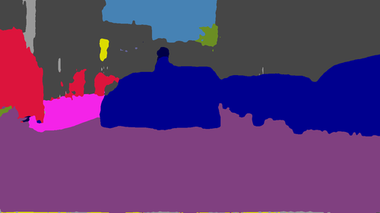} 
 			& \includegraphics[width=128px, valign=m]{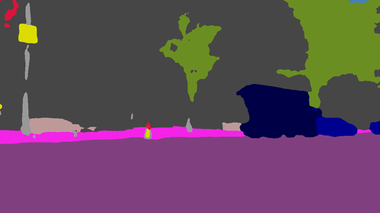} 
 			& \includegraphics[width=128px, valign=m]{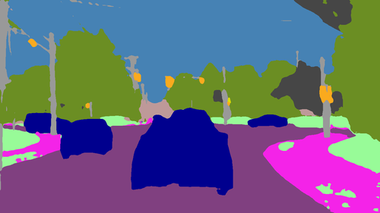} 
 			& \includegraphics[width=128px, valign=m]{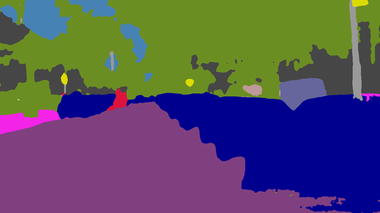}\vspace{0.3cm}\\
 			\hline\\
 			\noalign{\smallskip} \rotatebox{90}{w/o OMS \& WPL} 
 			& \includegraphics[width=128px, valign=m]{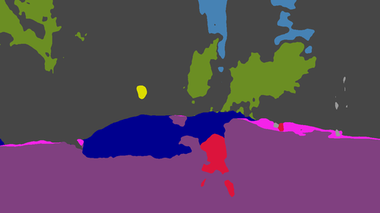} 
 			& \includegraphics[width=128px, valign=m]{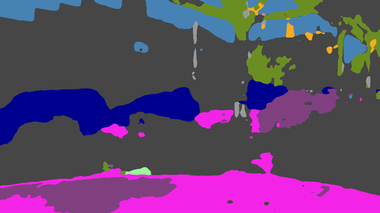} 
 			& \includegraphics[width=128px, valign=m]{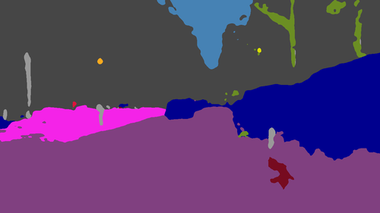} 
 			& \includegraphics[width=128px, valign=m]{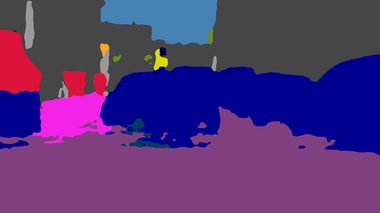} 
 			& \includegraphics[width=128px, valign=m]{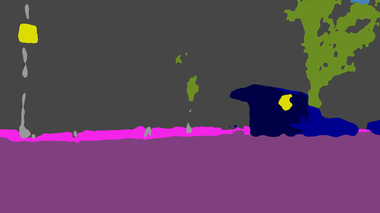} 
 			& \includegraphics[width=128px, valign=m]{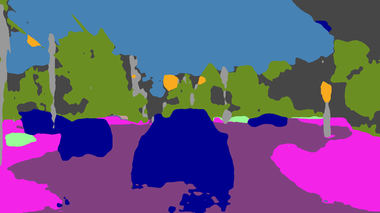} 
 			& \includegraphics[width=128px, valign=m]{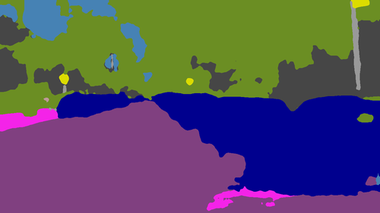}  \\
 			\noalign{\smallskip} \rotatebox{90}{w/o WPL} 
 			& \includegraphics[width=128px, valign=m]{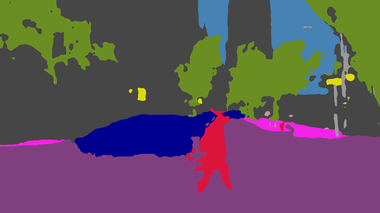} 
 			& \includegraphics[width=128px, valign=m]{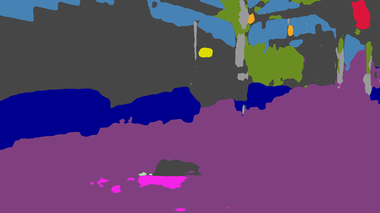} 
 			& \includegraphics[width=128px, valign=m]{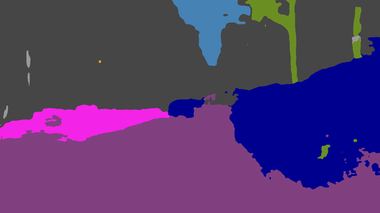} 
 			& \includegraphics[width=128px, valign=m]{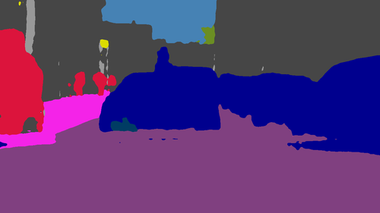} 
 			& \includegraphics[width=128px, valign=m]{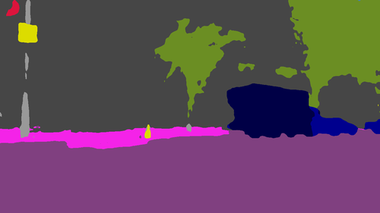} 
 			& \includegraphics[width=128px, valign=m]{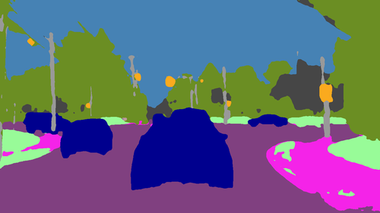} 
 			& \includegraphics[width=128px, valign=m]{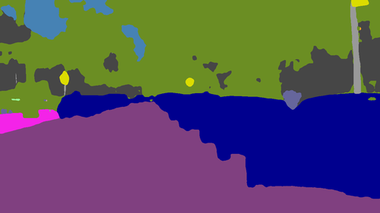}  \\
 			\noalign{\smallskip} \rotatebox{90}{Ours} 
 			& \includegraphics[width=128px, valign=m]{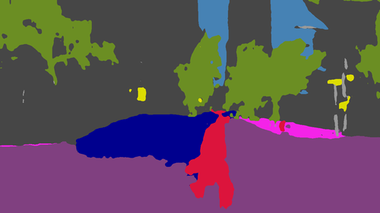} 
 			& \includegraphics[width=128px, valign=m]{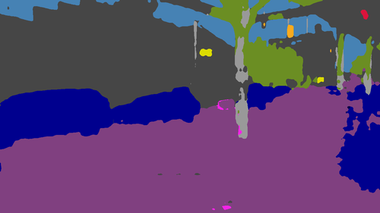} 
 			& \includegraphics[width=128px, valign=m]{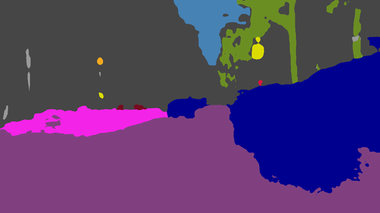} 
 			& \includegraphics[width=128px, valign=m]{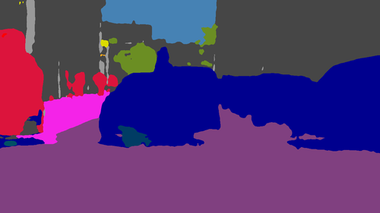} 
 			& \includegraphics[width=128px, valign=m]{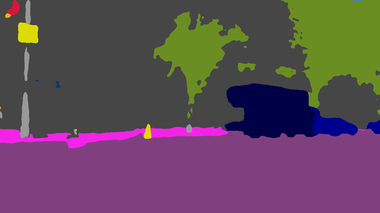} 
 			& \includegraphics[width=128px, valign=m]{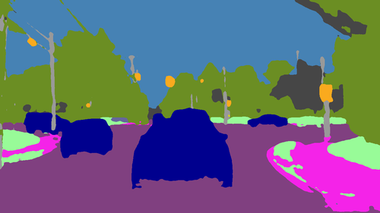} 
 			& \includegraphics[width=128px, valign=m]{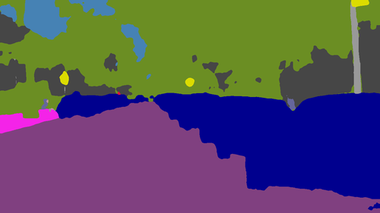}  \\
 			
 	\end{tabular}}
 	\caption{Comparison of our method with the state-of-the-art for semantic segmentation UDA.}\label{figure:sota}
 \end{figure*}
 \begin{table*}
 	\centering
 	\resizebox{\textwidth}{!}{\begin{tabular}{c|c|c|c|c|c|c|c|c|c|c|c|c|c|c|c|c|c|c|c|c}
 			{\textbf{Method}} &  {{\textbf{mIoU}}} & {{road}} & {{sidewalk}} & {{building}} & {{wall}} & {{fence}} & {{pole}} & {{t light}} & {{t sign}} &{{veg}} & {{terrain}} & {{sky}} & {{person}} &{{rider}} & {{car}} & {{truck}} & {{bus}} & {{train}} & {{m. bike}} & {{bike}} \\
 			\hline Baseline    &31.67     & 77.40 &39.95 &61.20 &12.01 &24.76 &23.68 &13.21 &24.11 &58.33 &27.18 &78.86 &24.73 &12.78 &63.34 &24.01 &28.43 &0.00 &4.76 &2.90 \\
 			\hline AdaptSegNet \cite{tsai2018learning}  &33.44   &  82.23 &39.85 &62.06 &9.84 &17.73 &20.39 &10.91 &22.47 &66.30 &22.81 &76.54 &32.24 &38.49 &68.95 &13.08 &30.31 &0.00 &17.97 &3.26 \\
 			\hline BDL \cite{li2019bidirectional} &39.60 & 83.18 &\textbf{48.78} & \textbf{73.93} & \textbf{30.87} & \textbf{27.33} &26.03 & \textbf{15.10} &26.05 &72.63 &26.08 &\textbf{88.01} &28.59 &26.59 &\textbf{76.37} &43.31 &\textbf{50.11} &0.00 &7.38 &2.10 \\
 			\hline\hline w/o OMS \& WPL &35.18 & 79.00 &37.23 &62.36 &8.60 &14.78 &20.98 &11.94 &22.92 &68.02 &13.11 &82.55 &38.96 &\textbf{44.61} &72.34 &29.10 &39.40 &0.00 &19.32 &3.16 \\
 			\hline w/o WPL            &  39.72 &            82.53 &44.51 &69.97 &20.29 &22.91 &\textbf{28.93} &14.02 &29.17 &74.32 &\textbf{28.98} & 83.53 &36.75 &32.80 &71.29 &43.03 &46.34 &0.00 &\textbf{21.80} &3.54   \\
 			\hline Ours       &            \textbf{40.04} & \textbf{84.03} &44.09 & 70.51 & 24.10 & 23.02 &28.31 & 14.08 &\textbf{30.07} &\textbf{75.31} &27.89 &83.49 &\textbf{39.10} &33.63 & 74.70 &\textbf{48.60} & 49.34 &0.00 &6.77 &\textbf{3.80}      \\
 	\end{tabular}}
 	\caption{State-of-the-art comparison. OMS refers to Online Multimodal Style-sampling. WPL is the Weighted Pseudo Labels strategy.}\label{table:sota}
 \end{table*}
 \begin{table}
 	\centering
 	\resizebox{0.5\linewidth}{!}{\begin{tabular}{c|c}
 			\textbf{Pseudo-labels} & \textbf{mIoU target} \\
 			\hline None & 39.77 \\			
 			\hline Batch-wise & 38.23 \\
 			\hline WPL (Ours) & \textbf{40.04} \\
 	\end{tabular}}
 	\caption{Comparison of various Pseudo Labeling strategies: Batch-wise, with our WPL, or with None.}\label{table:ablation2}
 \end{table}
\subsection{Bridged image-to-image translation}
\label{sec:exp_i2i}
We evaluate our bridged i2i (Sec.~\ref{sec:method_bridge}) on the Cityscapes to BDD-rainy translation task, and compare results against the recent CycleGAN~\cite{zhu2017unpaired} and MUNIT~\cite{zhu2017toward}. 
As stated, our i2i uses a MUNIT based and is referred to as MUNIT-bridged. 
It is trained on the bridged versions of the two datasets.
Training follow details from Sec.~\ref{paragraph:i2isettings}, except for CycleGAN that follows the original implementation\footnote{\cite{zhu2017unpaired} claims that best performances are obtained keeping constant the learning rate for half the training process (100k iterations in this case) to 2e-4 and then linearly decreasing to 0.}.\\
\noindent{}We argue - like others - that GAN metrics aren't appropriate for such evaluation. Thus, we report qualitative evaluation and segmentation task evaluation, together with usual GAN metrics.

\paragraph{Qualitative evaluation.} 
Fig.~\ref{figure:samefig} shows randomly selected samples from the Cityscapes validation set\footnote{For MUNIT and our method, MUNIT-bridged, we also randomize the style.}. 
It is visible that both CycleGAN and original MUNIT method fails at modeling the rain appearance, probably due to the large domain gap.
In detail, CycleGAN brings no realistic changes to the scene appearance, only adjusting color-levels in the image.
Original MUNIT, instead, seems to have collapsed and fails to produce significant outputs, probably due to instability related to the domain gap.
Conversely, our MUNIT-bridged model is the only one able to add realistic traits of rain in the synthetic images, thanks to the domain bridge. 

\paragraph{Quantitative evaluation.}
We compute GAN metrics following usual practices from~\cite{huang2018multimodal} and report results in Tab.~\ref{table:ganmetrics}.
The LPIPS distance~\cite{zhang2018unreasonable, shen2019towards} measures the image diversity~\cite{huang2018multimodal}, while the Inception Score evaluates both quality and diversity~\cite{salimans2016improved}.
In detail, LPIPS is the average on 19 paired translation of 100 images, and we report the diversity of real data in the target dataset as \textit{upper bound}. 
Inception Score uses the InceptionV3 network previously trained to classify source and target images.\\
\noindent{}Overall, we successfully improve performances over CycleGAN in both metrics, but original MUNIT has significantly higher performance. 
However, the images generated by MUNIT are evidently unrealistic (cf. Fig.~\ref{figure:samefig}) and thus we argue that GAN metrics are unreliable which is in fair alignment with~\cite{Barratt2018ANO} advocating that Inception Score is uncorrelated with image quality.

For a more comprehensive evaluation, we train a segmentation task on GAN translated clear$\rightarrow$rain images, and evaluate the standard mIoU metric on real rain images, reporting results in Tab.~\ref{table:semseg}. 
Note that for a fair comparison, we only sample a single style for MUNIT-based models, and report results when only trained on clear images as \textit{baseline}.
If the domain gap were reduced by the GAN translations, an improvement should be visible.\\
Instead, from the table, training on the original MUNIT-translated dataset leads to a significant performance decrease disproving the high GAN metrics.
Finally, our method outperforms CycleGAN by a little margin although CycleGAN fails to produce good quality images. 
Conversely, our method simultaneously reduces the domain shift and increases realism, thus it also eases performances evaluation on synthetic data.

\subsection{Unsupervised Domain Adaptation}
\label{sec:exp_UDA}

We now evaluate our UDA contributions encompassing our i2i translation methodology and compare with AdaptSegNet~\cite{tsai2018learning} and BDL~\cite{li2019bidirectional}, the best found recent works. We do not compare with less recent methods as CyCADA~\cite{hoffman2017cycada} since the approaches we are evaluating have already demonstrated to guarantee superior performances in UDA~\cite{tsai2018learning, li2019bidirectional}.
For fair comparison and given architectural similarities, BDL was adapted to work with the same segmentation network, data augmentation policy and hyperparameters detailed in Sec.~\ref{paragraph:segmentation}.
\\
Quantitative results are shown in Tab.~\ref{table:sota} where \textit{Ours} refer to the proposed UDA using domain-bridge i2i translation with Online Multimodal Style-sampling (OMS) and Weighted Pseudo Label (WPL), and \textit{w/o WPL} or \textit{w/o OMS \& WPL} the self-explanatory ablation versions.
\textit{Baseline} refers to the training without any UDA.
Overall, our methodology performs on par~($+0.44$) with BDL, the best state-of-the-art, using a much simpler domain adaptation method, and significantly better~($+6.6$) than AdaptSegNet. 
Studying the contributions of our OMS and WPL contributions, all components are necessary to reach the best performances. 
Qualitative evaluation on the target dataset is shown in Fig.~\ref{figure:sota}, and in fair alignment with quantitative metrics.

\paragraph{Weighted Pseudo Labels.}
We evaluate the effectiveness of our WPL proposal and report results in~Tab.~\ref{table:ablation2}, comparing similar training with either \textit{WPL} (Ours), \textit{batch-wise} Pseudo-Label\footnote{For batch-wise Pseudo-Label implementation, we compute optimal threshold per class and per batch.}, or \textit{None}.
For all, the training is performed using as target the whole BDD100k train set (removing duplicates from 10k split) together with the rainy sequences from Domain-bridge dataset, resulting in over 90k images. Performance is reported on target BDD-rainy.
We do not compare against offline Pseudo Label, as this would be impractical with such a big dataset, and this evaluation is partly encompassed in BDL comparison (cf. Tab.~\ref{table:sota}).
For WPL, we empirically set $\gamma = 1$, $\sigma = 0.005$ (Eq.~\ref{loss-ss}) to balance contributions and $p_{tp}=0.75$ (Eq.~\ref{eq:UDA_loss_source} \& \ref{eq:UDA_loss_target}). 

From results in Tab.~\ref{table:ablation2}, \textit{WPL} performs the best and \textit{batch-wise} Pseudo Label third.
In fact, the performance decrease for \textit{batch-wise} (compared to no self-supervision) may be explained since best batch pixels are used as pseudo-labels, thus possibly implying some incorrect self-supervision in case of low batch accuracy.
Instead, our \textit{WPL} boosts the mIoU on the target dataset which is expected due to its expansion behavior. 
\section{Conclusions}
In this work, we introduced a novel approach to generate realistic rainy images with an i2i network, while preserving traits of adverse weather that are typically ignored by state-of-the-art architectures. Then, we extended our system and demonstrated its performances in UDA for semantic segmentation, and with a simple pipeline we obtained on par performances with respect to the state-of-the-art. Finally, we introduced a novel pseudo labeling strategy that works with an unlimited number of images, and has an optimizable weight parameter used to guide region growing. For future work, we plan to extend our pseudo labeling approach with class-wise thresholds, and to extend the method to different adverse weather conditions.

{\small
	\bibliographystyle{ieee}
	\bibliography{egbib}
}
	
\end{document}